 \documentclass[final,1p,times]{elsarticle}

\usepackage{caption}
\renewcommand\figurename{Fig}
\usepackage{pdflscape}
\usepackage{amsthm}
\theoremstyle{plain}
\theoremstyle{definition}

\usepackage{afterpage}

\usepackage{graphicx}
\usepackage{epstopdf}

\usepackage{amssymb}
\begin{document}
\begin{frontmatter}



\title{A brief network analysis of Artificial Intelligence publication}


\author[address1]{Yunpeng Li}
\author[address1]{Jie Liu}
\author[address1,address2]{Yong Deng\corref{label1}}

\cortext[label1]{Corresponding author: Yong Deng, School of
Computer and Information Science, Southwest University, Chongqing,
400715, China. Email address: ydeng@swu.edu.cn,
prof.deng@hotmail.com. Tel/Fax:(86-23)68254555.}

\address[address1]{School of Computer and Information Science, Southwest University, Chongqing, 400715, China}
\address[address2]{School of Engineering, Vanderbilt University, Nashville, TN, 37235, USA}
\begin{abstract}
In this paper, we present an illustration to the history of Artificial Intelligence(AI) with a statistical analysis of publish since 1940. We collected and mined through the IEEE publish data base to analysis the geological and chronological variance of the activeness of research in AI. The connections between different institutes are showed. The result shows that the leading community of AI research are mainly in the USA, China, the Europe and Japan. The key institutes, authors and the research hotspots are revealed. It is found that the research institutes in the fields like \textit{Data Mining}, \textit{Computer Vision}, \textit{Pattern Recognition} and some other fields of \textit{Machine Learning} are quite consistent, implying a strong interaction between the community of each field. It is also showed that the research of \textit{Electronic Engineering} and Industrial or Commercial applications are very active in California. Japan is also publishing a lot of papers in robotics. Due to the limitation of data source, the result might be overly influenced by the number of published articles, which is to our best improved by applying network keynode analysis on the research community instead of merely count the number of publish.
\end{abstract}

\begin{keyword}

Artificial Intelligence \sep Network Analysis \sep Keyword Network
\end{keyword}

\end{frontmatter}


\section{Introduction}\label{sec:intro}

Artificial Intelligence has been a long-pursuing goal of scientists, technicians and even novelists and philosophers long before the birth of electronic computer\cite{buchanan2005very,haugeland1989artificial,churchland2012could,anderson1986machine,111112}. But it is until the 1940s when electronic computing machines was successfully developed that we have a chance to create such fantasy in places outside our dreams and papers\cite{buchanan2005very,kitano2007intelligence,1111123}. The invention of computer becomes a trigger in the history of AI, booming the experimental research which in turn prompts the theoretical developments as well\cite{marr1977artificial,11111}. Over the past 73 years, many kinds methods has been raised, developed and diminished, as well as communities and researchers.

A lot of reviews \cite{4804025,gregory1982heroes,}and short histories\cite{111112} has published to give brief summary of the development of AI. Most of such reviews are made by the witness of the big things\cite{buchanan2005very}. In this paper, we present an illustrative review of the history of Artificial Intelligence(AI) by statistically analysing the publish over the past 73 years. Many resources of online publish data base has been considered before choosing the \textit{IEEE Xplore} as the unique data recourse of the research. 610,051 articles are returned responding to our query, which occupies 1/6 of the entire IEEE database. Then we applied natural language processing techniques to analysis the meta data returned by the query. The authors, affiliations and keywords are the main fields in our analysis. Then we build a coauthor network of authors as the fundament of our analysis. The coreness of each author is caculated (to be added). The importance in research of each institute and country is based on the authors' coreness in them(to be added).

By clustering the keywords network which is derived from the co-occurring relations of the keywords, a keyword-to-field table is made as the fundament of the by-field analysis. It is showed that while the research in \textit{Data Mining}, \textit{Pattern Recognition}, \textit{Computer Vision} and \textit{Machine Learning} are quite consistent, \textit{Electronic Engineering}, \textit{Robots} and application-orientated researches are highly centralized. 

\section{The method}\label{sec:method}

The data applied in this research is retrieved via \textit{IEEE Xplore} XML search API at\cite{WinNT,cve}. We choose IEEE API because it is sufficiently representative, abundant in volume and convenient for use. Most of other scholar APIs we have also referred to can be found at http://libguides.mit.edu/apis. The raw data was allocated with a query word of "AI", "\textit{Data Mining}", "Natural Language Processing" and synonyms in the field of title and keywords (including Thesaurus Terms, Inspect Controlled Terms and Index Terms, see definitions at http://ieeexplore.ieee.org/gateway/). As of Oct. 15, 2013, the query returned 610,051 results of the 3,563,516 articles in IEEE data base. The total volume of the XML reply files is 1.39 GB.

Then the raw data is processed with R. Unnecessary symbols like quotes, slashes and dashes are all replaced with blanks. For the scale of consistency, all the "and" are replaced with symbol \&. All the "the", no matter leading or not, are eliminated. All the "at" are also eliminated because they are often placed between the name of a university and the location of the campus, and might be left out in other cases which causes duplication. Leading, tailing and duplicated blanks are 
eliminated. All abbreviations and acronyms occurring 10 times or above are rewritten, like, among which the most popular abbreviation "Univ.", is rewritten as "University". Some synonyms are also unified. For instance, "University of California" are replaced with "California University".

Besides structuralizing and calibrating the meta data, we also applied some text mining processes to excavate the affiliation (university, institute or company) and geological information (city, state, postcode and nation) of each author for further analysis. Since each paper is related to one or more authors but only one affiliation, we associated this affiliation information with the first author. Then we aggregated the geological information of all the authors in certain organization to get the geological information of the organization itself. Typically the needed information is the most frequent non-empty string of such field. However ,in order to deal with organizations with multiple sites (typically 
universities with different campuses like California University and New York University, which has campuses in different countries), all the cities meeting certain conditions need to be reserved. In this study, to aggregate the city information, we kept the most frequent non-empty one, and all others of which the frequency is beyond 5 times and 3 percent of the most frequent one at the same time. Then the state, postcode and nation terms are the most frequent non-empty one of each field in each city. We then adhere the city name at the end of the organization name for secernment. In following processes, different sites are regarded as independent entities.

Further more, most organizations are then accompanied with their coordinate in order to illustrate some of our results with a map. For 2105 organizations in the US, all coordinates are queried with google map geocode API. Then 27,715 organizations outside US are matched with a coordinate dataset of 257,495 cities all over the world, nation by nation. After these processes, 14,422 organizations are still not matched with coordinates. Then we tried to match these organizations with our dataset of cities to the name of the organization. 1350 institutes are such matched. Then we processed the rest again with google map API. At this step, 6,546 out of all the 29,820 organizations are matched with its coordinate. Then we reverse geocode the coordinates to check for validity and to complete nation / state fields. 351 coordinates are found inconsistent with nations, most of them (35) are at French-Swiss border. We checked them with google map and believe such error is due to the minor mistakes of map at the border. These coordinates are reserved and all others are taken as mistakes and reset to NA.


\section{Results}\label{sec:overall}

The processed data is then analyzed for revealing trends and connections in AI research. We first build the network of authors with coauthor relationship. For each paper, we build a undirected complete graph of weight 1 between all authors. The links connecting the same pair of nodes are then accumulated. In total, there are 4,954,982 links between 662,762 authors with a total weight of 7,133,757. The distribution of degrees of each author is as following:

\begin{figure}
    \centerline{\includegraphics[scale=0.4]{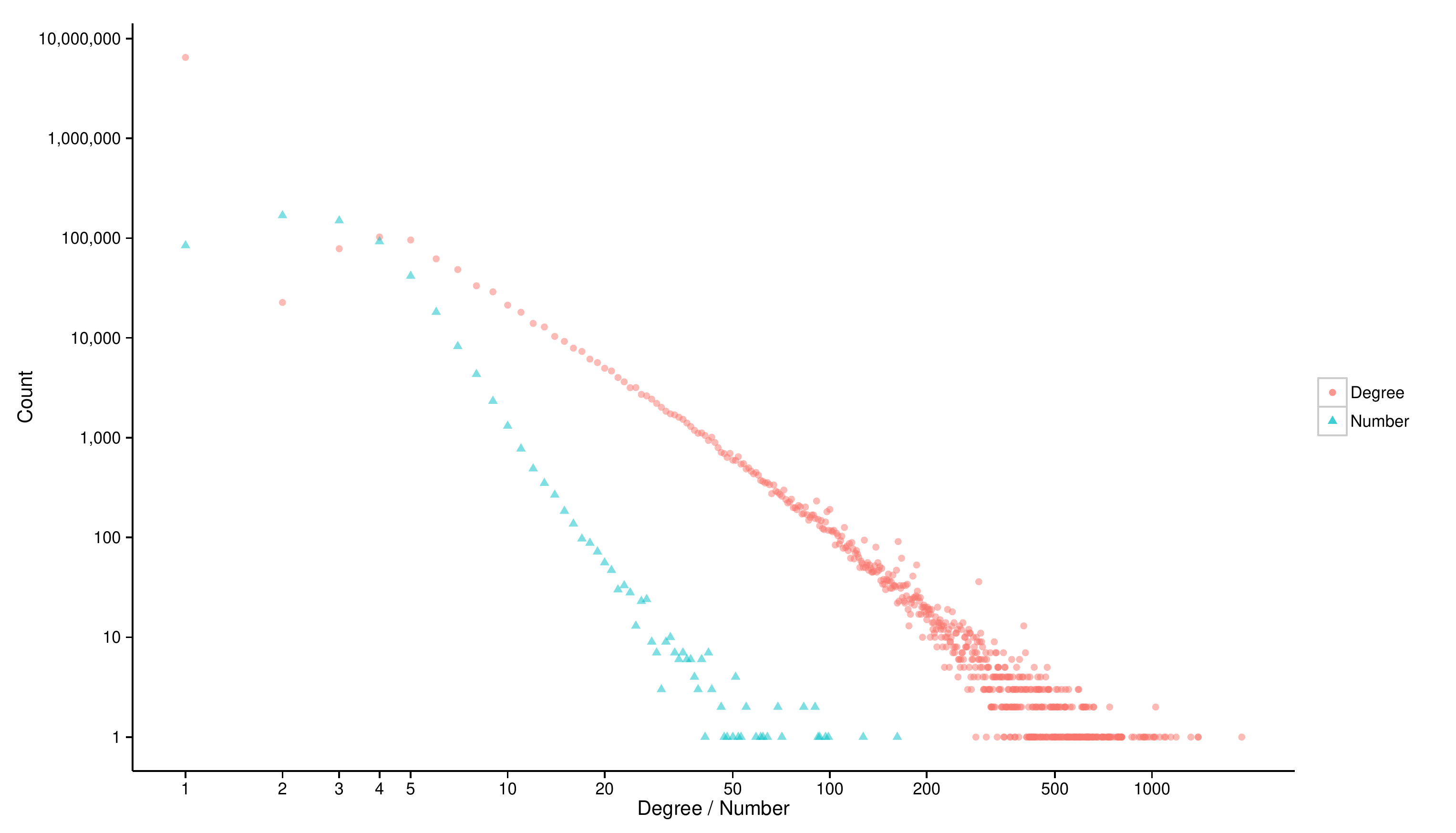}}
\caption{Degree distribution of coauthor network of AI research(o) and distribution of author numbers of papers ($\Delta$) on a double logarithmic scale}
\label{fig:author_dstr}       
\end{figure}

It can be seen from Fig. \ref{fig:author_dstr} that the coauthor network fits the power-law quite well. This suggests that the coauthor network is possibly a scale-free complex network.

It is also showed in Fig. \ref{fig:author_dstr} an abnormal decrease in number of authors with degree of 2 to 4. The reason might be that most of the papers are published in a group of 3 rather than 1 or 2. It could be seen that the number of authors in each paper also have a decrease at the beginning, which might be the reason. In this we may conclude that most of research in AI are made up of three or more participants.

\subsection{Global connection}\label{sec:connect}

After getting the coauthor network, we used it to find the cooperation relations between organizations, mainly universities. Generally, Fig. \ref{fig:global_connction} maps the coauthor network to the globe. Each line is the geological big circle (\textit{i.e.} the shortest path between two points on the sphere) of the linked organizations. From the 4,954,982 coauthor relations between authors, we obtained 486,120 connections between 29,820 organizations. The connections are all illustrated on Fig. \ref{fig:global_connction}.

\afterpage{%
\thispagestyle{empty}
    \begin{landscape}
        \begin{figure}
        \vspace*{-4cm}
            \centerline{\includegraphics[scale=0.6]{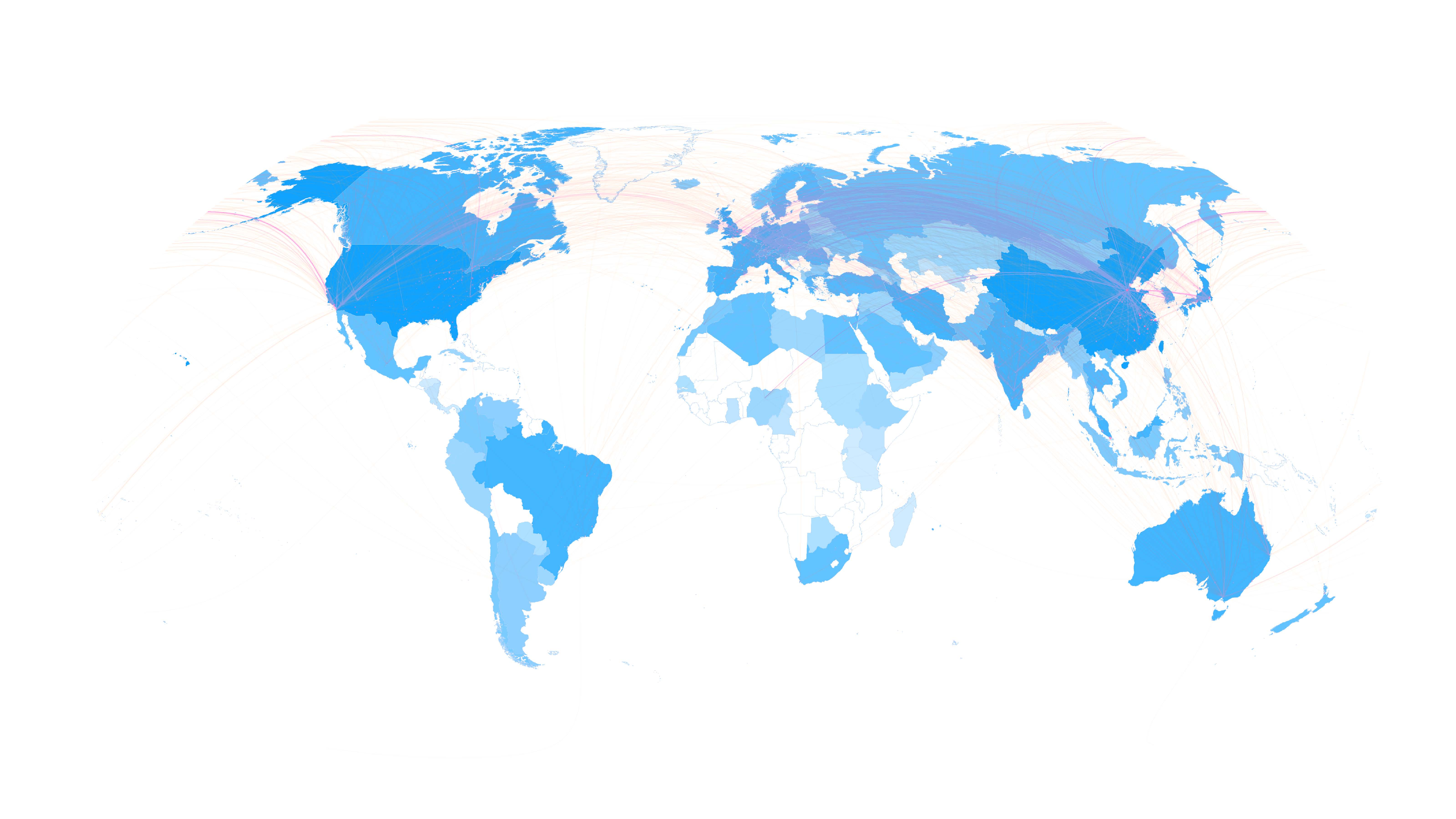}}
        \caption{Global connection of universities, institutes, companies and other research institutes}
        \label{fig:global_connction}       
        \end{figure}
    \end{landscape}
\clearpage
}

It could be clearly seen on the map that most of the researchers locate in the United States of America, China, Japan, and some countries of Europe like the United Kingdom, Germany and France. Researches in Brazil, Singapore, India and Australia are also very active. These national differences are more obvious in Fig. \ref{fig:global_connction_nation}, which accumulates the links by nation. It is significant that the United States and China are leading partners of global research cooperations. Japan, although geologically close to China, are more connected to the US and European countries academically.

\afterpage{%
\thispagestyle{empty}
    \begin{landscape}
        \begin{figure}
        \vspace*{-4cm}
            \centerline{\includegraphics[scale=0.6]{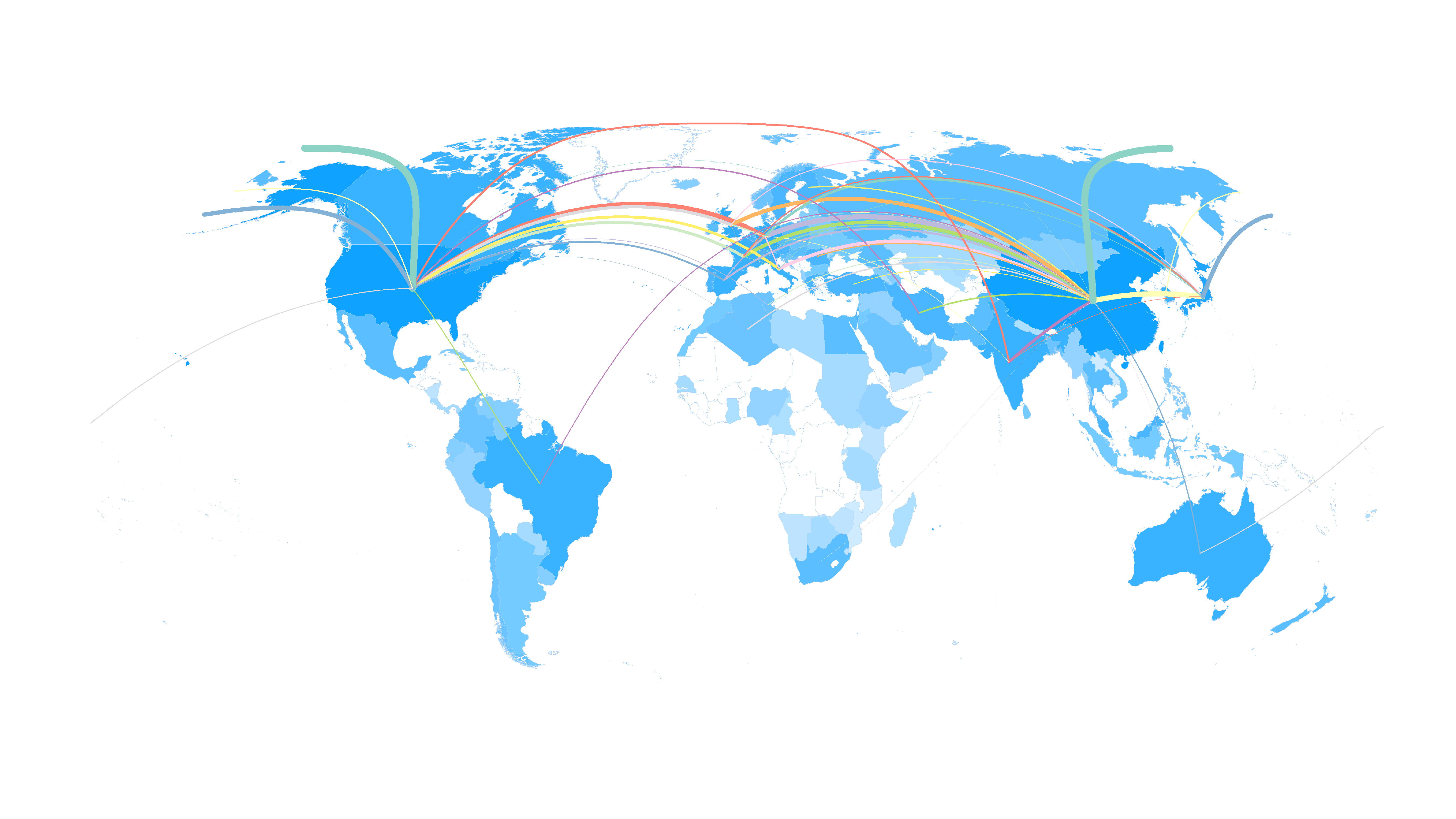}}
        \caption{Global connection of universities, institutes, companies and other research institutes}
        \label{fig:global_connction_nation}       
        \end{figure}
    \end{landscape}
\clearpage
}

\subsection{Global distribution}\label{sec:distri}



With the method to be described in \ref{sec:term_field}, we also get maps of activity in each field of AI. It could be seen that the research institutes in the fields like \textit{Data Mining}, \textit{Computer Vision}, \textit{Pattern Recognition} and some other fields of \textit{Machine Learning} are quite consistent, implying a strong interaction between the community of each field. The most active research has taken places in the institutes of the United States, China, Singapore and the United Kingdom. Despite the consistency, yet a small but interesting difference in such fields is that the activeness in \textit{Computer Vision}, \textit{Pattern Recognition}, \textit{Data Mining} are in descending order for the west countries (the United States and England), but ascending for the east countries (China and Singapore). It is also clear that the United States is extremely competitive in the research of \textit{Electronic Engineering}, especially for the Californian Institutes, which is also very strong at combining AI with industrial and commercial products. As for \textit{Robots}, the United States, especially the northeast corner of which, and Japan are dominating most of the researches, which is not unexpected at all.

\afterpage{%
\thispagestyle{empty}
    \begin{table}[!htb]
        \centering
        \vspace*{-4cm}
        \hspace*{-4cm}
        \begin{tabular}{cc}
            \includegraphics[scale=0.025]{DM.pdf}
            &
            \includegraphics[scale=0.025]{CV.pdf}
            \\
            (a) Data Mining
            &
            (b) Computer Vision
            \\
            \includegraphics[scale=0.025]{PR.pdf}
            &
            \includegraphics[scale=0.025]{ML.pdf}
            \\
            (c) Pattern Recognition
            &
            (d) other fields of Machine Learning
            \\
            \includegraphics[scale=0.025]{robot.pdf}
            &
            \includegraphics[scale=0.025]{EE.pdf}
            \\
            (e) Robot
            &
            (f) Electronic Engineering
            \\
            \includegraphics[scale=0.025]{CS.pdf}
            &
            \includegraphics[scale=0.025]{application.pdf}
            \\
            (g) other fields of Computer Science
            &
            (h) Industrial and Commercial Applications
        \end{tabular}
        \caption{global research interest distribution in AI}
    \end{table}
\clearpage
}

\begin{figure}[!htb]
    \centerline{\includegraphics[scale=1]{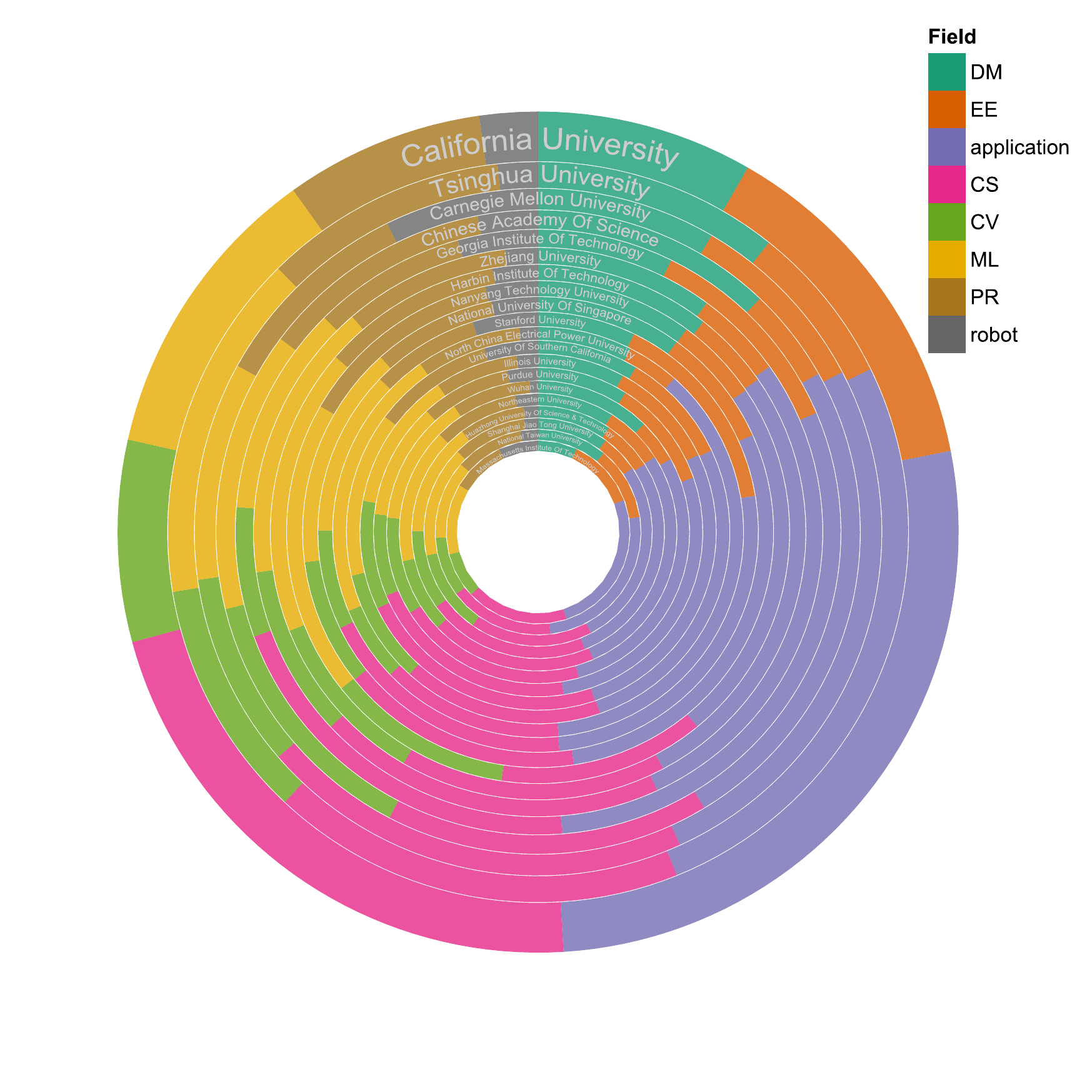}}
\caption{Top researching universities in AI}
\label{fig:top_univ}       
\end{figure}

It is also interesting to statistic the performance of the top researching institutes, mainly universities, and their area of professional. The 20 most active universities in the community of AI is showed in Fig. \ref{fig:top_univ}. It is clearly seen that the California University is the biggest host to many productive researchers. Best of our effort has been made to clarify the affiliation of them but due to the incompleted addresses left by many of the researchers, the exact affiliation of many of them are still indistinguishable, leaving the California University the biggest host to researchers. As for followers, Tsinghua University is the most productive institute in AI, if we choose not to take the California University as a single entity. Tsinghua is competitive in almost every field of AI, only to except \textit{Robots} and EE, which are dominated by the United States and Japan. On the other hand, Carnegie Mellon University, the second productive university in AI, is extremely competitive in \textit{Robots}, which, in our statistics, is even stronger than California University as a whole. Massachusetts Institute of Technology, which ranks the 20th in this list, might be suffering an under-representation from the number of the publish.

It is also found that the groups of research team are generally much bigger in China and Singapore than in the US etc. Explanations and analysis to be added in the next version.

\subsection{From keywords to fields}\label{sec:term_field}

In order to analysis the presentation of entities in different fields, we need a table to map each keyword to a certain subfield of AI. There are some ways to accomplish such procedure like ontology, clustering or just assign the field manually. In this review, we made the classification via clustering and then manually checked the result.

\begin{figure}[!htb]
    \centerline{\includegraphics[scale=0.25]{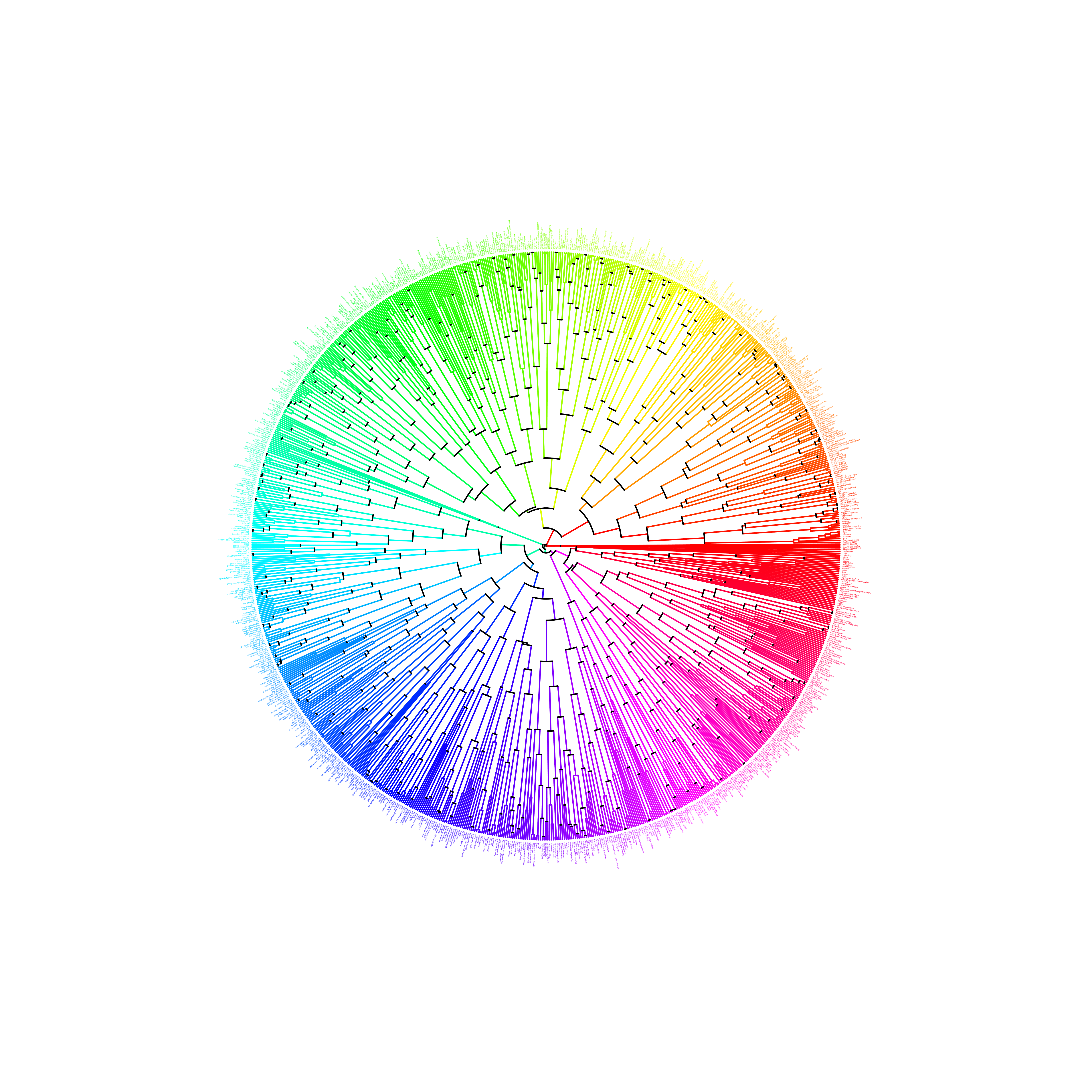}}
\caption{Global connection of universities, institutes, companies and other research institutes}
\label{fig:dend}       
\end{figure}

We monitor the most popular 1000 keywords, and build a network of their correlations. Each of the keywords of the same paper are linked with an undirected edge of weight 1, then duplicated edges are accumulated. After building the keyword network, which occupies 1000 keywords and 542,048 edges with a total weight of 3,716,812, we used the random walk algorithm to detect the community structure of the network. The hierarchical structure of the monitored keywords is showed in Fig. \ref{fig:dend}. Noting that a complete graph of 1000 nodes contains 999,000 edges, the keyword network contains 54.26\% of the complete graph, and holds an average weight of 6.85, implying quite a strongly connected graph.

Then we cut the dendrogram at a place to get 50 groups. The groups are manually classified into 8 fields of \textit{Data Mining} (DM), \textit{Computer Vision} (CV), \textit{Pattern Recognition} (PR, which includes keywords like handwriting recognition which involves both PR and CV), other fields of \textit{Machine Learning} (ML, mainly general proposed algorithms like Neural Networks, and supporting theories like rough set), \textit{Robots} (Robot), \textit{Electronic Engineering} (EE, mainly micro-electronics and communication), other fields of Computer Science (CS, mainly computer architecture related and programming methodology), and \textit{Application}+..3655555555555555555555555555555502042555555555555555555525 areas(application, including industrial and commercial products). The group assignment of 65 keywords out of the 1000 are manually corrected.

The monitored 1000 keywords are then applied to the dataset to find the main research area of each author. The activity of each author in certain field is related to the number of published articles in the field. Each paper is assigned to the 8 fields proportionally to its membership of the fields according to its keywords. For example, if an article has 10 keywords, 3 of them are in PR, 2 of the rest are in DM, and the rest are not monitored, then we say the article is 0.6 PR and 0.4 DM, which will contribute to the activity in these fields of the author. The monitored keywords occupy 20.67 \% ( 1,400,000 out of 6,771,660) of the keywords in the dataset. The activity of each author is the sum of the fields of all papers the author have ever published.The most productive authors in each field is listed in Fig. \ref{fig:field_author}(to be added).


With similar processes, we may also find the activity of research by field of institutes, cities and countries, as is already showed in Section \ref{sec:connect} and \ref{sec:distri}.

\subsection{Hot words}\label{sec:word_cloud}

With the keyword to field correspondence obtained in Section \ref{sec:term_field}, we made a further exploration into the hot words in recent 5 years. The 5 year span covers the publish in 2009-2013. As this research is carried out at mid-October, some of the publish in year 2013 might be unavailable at the time our data was collected. The result is showed in Fig. \ref{fig:hot_word}. It is clearly showed that technique-neutral words are generally more popular. "Data Mining","Learning", "Training", "Computational Modeling" are all among the hottest 10 keywords. "Support Vector Machines", as a technique-specific term, seems to be an exception, which is the $3^{rd}$ hottest keyword. "Feature Extraction", the $2^{nd}$ hottest one, benefiting from the rising attention in \textit{Computer Vision} and \textit{Pattern Recognition}, is neither as neutral as DM nor as specific as SVM. Meanwhile the "Educational Institutions" may be a dark horse for the MOOC movement is in full swing.

\begin{figure}
    \centerline{\includegraphics[scale=0.4]{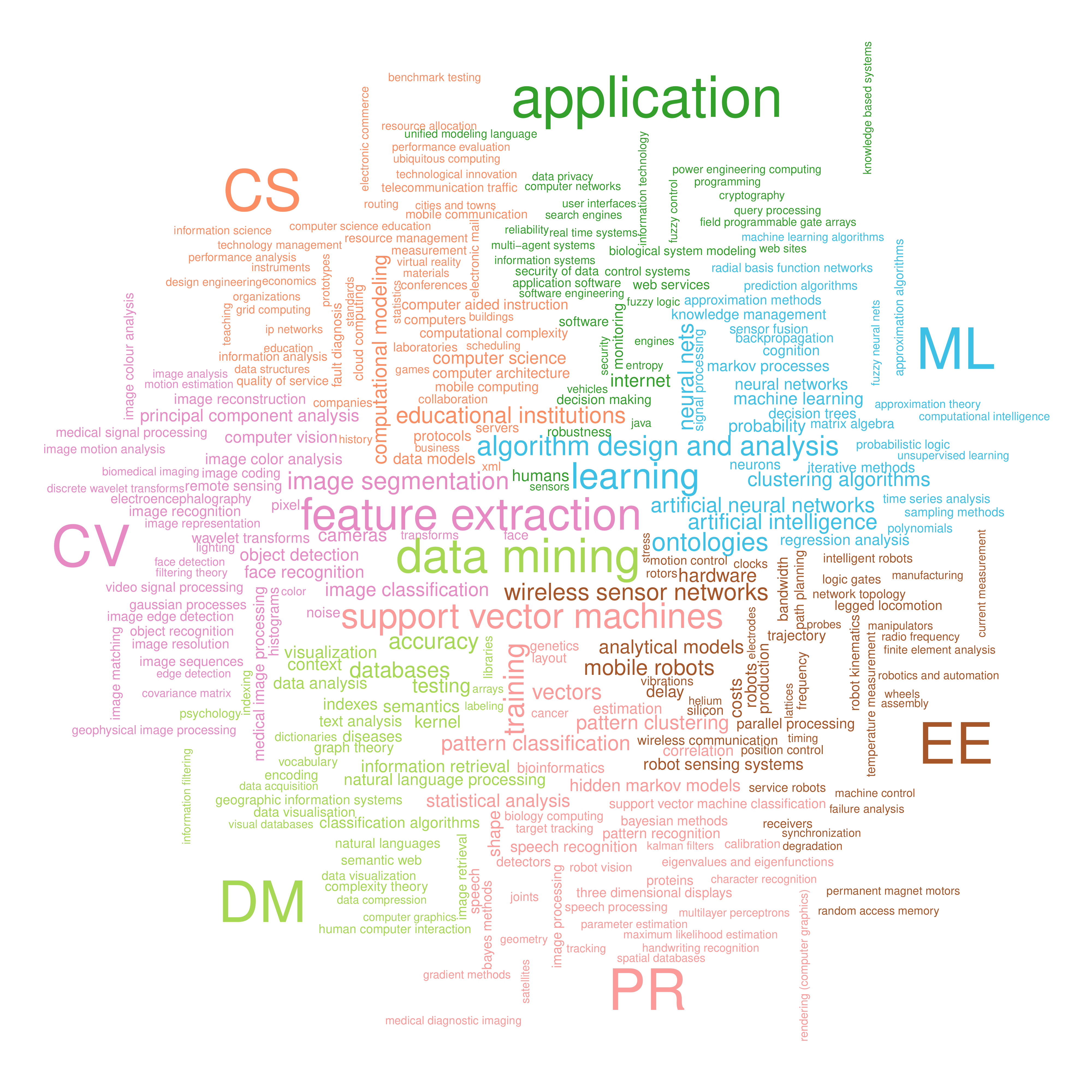}}
\caption{Hot words of recent 5 years' publish (2009-2013)}
\label{fig:hot_word}       
\end{figure}

\section{Chronological analysis}\label{sec:chrono}

Besides the features discussed in section \ref{sec:overall}, there might be some more inspiring information beclouded. While static analysis offers us some overall status of AI research, or what the status is like, chronological analysis can tell us how it becomes like this. In this section, we divided the dataset of papers by the year it was published, and reapplied some of the analysis carried out in \ref{sec:overall} to find out how the AI community evolves over the past 73 years (1940-2013).

\subsection{Total publish over year}\label{sec:annual_total}

Since the development of ENIAC, EDVAC and Colossus in the 1940s, Artificial Intelligence has come to more reality than just fantasy. Computers provided a chance to bring the theoretical possibilities of intelligent machines.

But the annual publish in AI seems not much affected by the events. The publish number raise exponentially steadily. The annual publish can be fitted with $publish=1.156*exp(year - 1935.596)$ with a 
Adjusted R-squared of 0.9841.

\afterpage{%
\thispagestyle{empty}
    \begin{landscape}
        \begin{figure}
            \centerline{\includegraphics[scale=0.6]{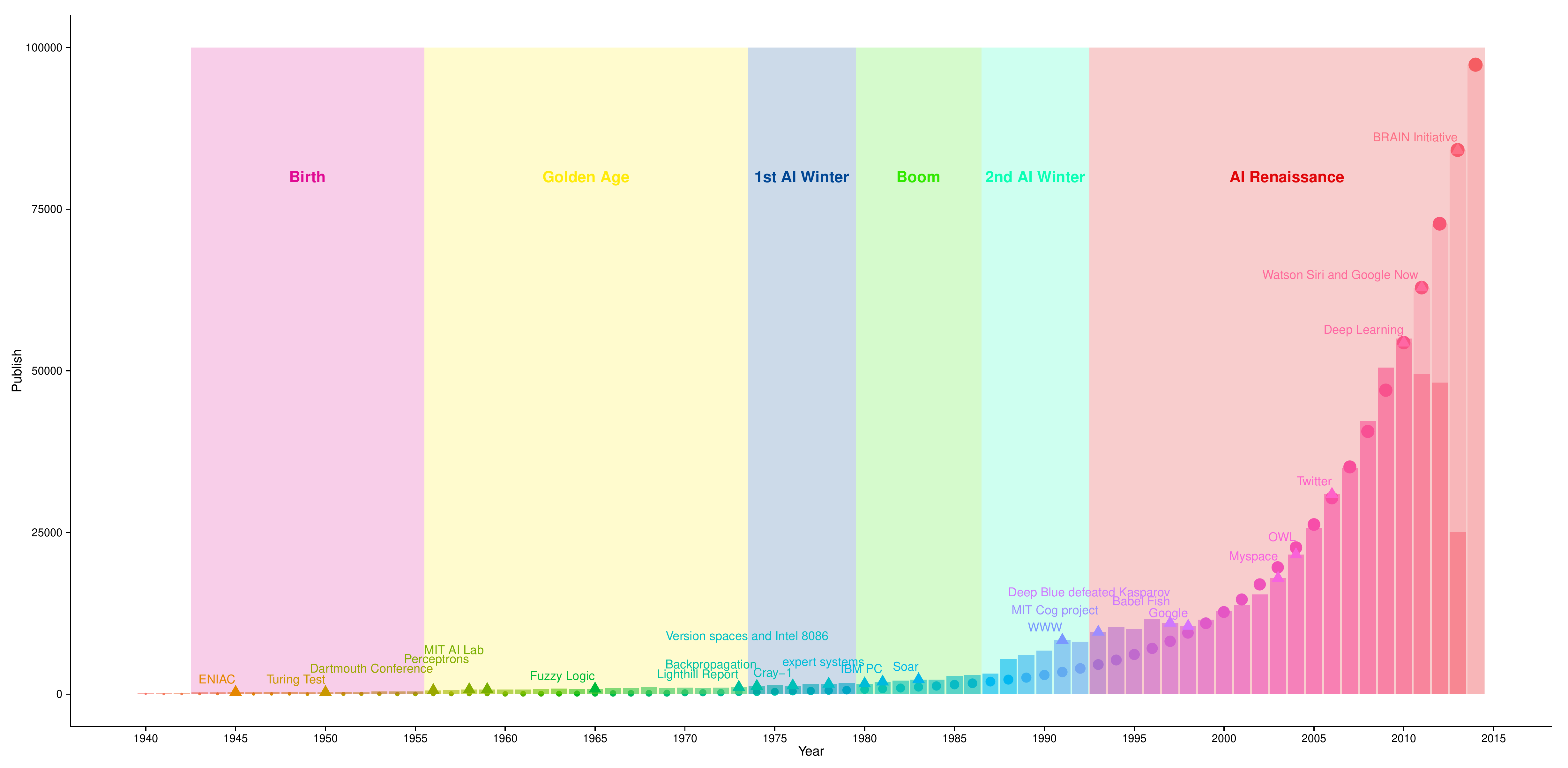}}
        \caption{Trend of annual publish since 1940}
        \label{fig:hot_word}       
        \end{figure}
    \end{landscape}
\clearpage
}
\subsection{Trend in keywords}\label{sec:annual_keyword}
%
%
%
%
\subsection{Community evolution}\label{sec:annual_community}

\section{Conclusion and outlook}\label{sec:conclusion}
%

In this review, a statistical review of the history of AI in publish is presented. The publish in IEEE data base is analyzed to give some holistic analysis of the global research in AI. We collected and mined through the IEEE publish data base to analysis the geological and chronological variance of the activeness of research in AI. The connections between different institutes are showed. The result shows that the leading community of AI research are mainly in the USA, China, the Europe and Japan. The key institutes, authors and the research hotspots are revealed. It is found that the research institutes in the fields like \textit{Data Mining}, \textit{Computer Vision}, \textit{Pattern Recognition} and some other fields of \textit{Machine Learning} are quite consistent, implying a strong interaction between the community of each field. It is also showed that the research of \textit{Electronic Engineering} and Industrial or Commercial applications are very active in California. Japan is also publishing a lot of papers in robotics. Due to the limitation of data source, the result might be overly influenced by the number of published articles, which is to our best improved by applying network keynode analysis on the research community instead of merely count the number of publish.

In this paper, due to the limitation of data source, we cannot distinguish the papers that are harbingers and milestones from the mediocrities which are only imitators or parody. The quantity of publish outweighed the quality. The innovativeness, inspiration and impact are the key values of a paper, which varies much between mileposts and imitators. In our review, due to the lack of evaluation criteria, each paper is considered equal in significance. The best effort of us has been put forward to address this awkward blemish. We performed our analysis mainly of the network property when trying to identify influential entities. An improvement could still be made if a citation map of papers is available. One may assign a significant value of papers due to their citation, or pagerank. This may improve the current situation in which quantity of publish is too much valued, which is also overwhelming in some parts of the world.

\section*{Acknowledgment}

The work is partially supported by National Natural Science Foundation of China (Grant No. 61174022), Chongqing Natural Science Foundation (Grant No. CSCT, 2010BA2003), Program for New Century Excellent Talents in University (Grant No. NCET-08-0345), Doctor Funding of Southwest University (Grant No. SWU110021).





\bibliographystyle{elsarticle-num-names}
\bibliography{AI_Network_ref}







\end{document}